\documentclass{article}

\PassOptionsToPackage{square,sort,comma,numbers}{natbib}
\usepackage[final]{neurips_2019_ml4ad}

\usepackage[utf8]{inputenc} 
\usepackage[T1]{fontenc}    
\usepackage{hyperref}       
\usepackage{url}            
\usepackage{booktabs}       
\usepackage{amsfonts}       
\usepackage{nicefrac}       
\usepackage{microtype}      

\usepackage{graphicx}
\usepackage{algorithm}
\usepackage{amsmath}
\usepackage[noend]{algpseudocode}
\usepackage{color,soul}

\setcitestyle{square}

\title{Quadratic Q-network for Learning Continuous Control for Autonomous Vehicles}

\author{%
  Pin Wang$^{1}$\thanks{Corresponding author}, Hanhan Li$^{2}$, Ching-Yao Chan$^{1}$ \\
  $^{1}$ University of California, Berkeley\\
  $^{2}$ Google AI \\
  \texttt{\{pin\_wang, cychan\}@berkeley.edu} \\
  \texttt{uniqueness@google.com}
}

\begin{document}

\maketitle

\begin{abstract}
  Reinforcement Learning algorithms have recently been proposed to learn time-sequential control policies in the field of autonomous driving. Direct applications of Reinforcement Learning algorithms with discrete action space will yield unsatisfactory results at the operational level of driving where continuous control actions are actually required. In addition, the design of neural networks often fails to incorporate the domain knowledge of the targeting problem such as the classical control theories in our case. In this paper, we propose a hybrid model by combining Q-learning and classic PID (Proportion Integration Differentiation) controller for handling continuous vehicle control problems under dynamic driving environment. Particularly, instead of using a big neural network as Q-function approximation, we design a Quadratic Q-function over actions with multiple simple neural networks for finding optimal values within a continuous space. We also build an action network based on the domain knowledge of the control mechanism of a PID controller to guide the agent to explore optimal actions more efficiently. We test our proposed approach in simulation under two common but challenging driving situations, the lane change scenario and ramp merge scenario. Results show that the autonomous vehicle agent can successfully learn a smooth and efficient driving behavior in both situations.
\end{abstract}

\section{Introduction}

Reinforcement Learning (RL) has been applied in robotics for decades and has gained popularity due to the development in deep learning. In recent studies, it has been applied for learning 3D locomotion tasks (e.g. bipedal locomotion and quadrupedal locomotion \cite{schulman2015high}) as well as robot arm manipulation tasks (e.g. stacking blocks \cite{haarnoja2018composable}). Google DeepMind also showed the power of RL for learning to play Atari 2600 games \cite{mnih2013playing} and Go games \cite{silver2016mastering}. In these applications, either the operation environment is simple, e.g. no interaction with other agents, or the action space is limited to discrete selection, e.g. left, right, forward, and backward.

In regard to the driving tasks for autonomous vehicles, the situation is totally different because vehicles are required to operate smoothly and efficiently in a dynamic and complicated driving environment. Tough challenges frequently arise in driving domains. For example, the vehicle agent needs to coordinate with surrounding vehicles so as not to disturb the traffic flow significantly when it executes a maneuver, e.g. merging into a joint traffic flow. More importantly, the control action should be continuous to guarantee smooth travelling.  

There have been some efforts in applying RL to autonomous driving \cite{yu2016deep}\cite{ngai2007automated}\cite{sallab2016end}, however, in some of the applications the state space or action space are arbitrarily discretized (e.g. vehicle acceleration is split into some fixed values), to fit into the RL algorithms (e.g. Q-learning) without considering the specific features of the driving problem. This simplified discretization results in the loss of a complete representation of the continuous space. Policy gradient-based methods are alternatives for continuous action problems, but they often complicates the training process by involving a policy network and sometimes suffer from vanishing or exploding gradient problems.

In this study, we resort to model-free Q-learning and design the Q-function in a Quadratic form based on the idea of Normalized Advantage Function \cite{gu2016continuous}. With this form, the optimal solution can be obtained in a closed form. Additionally, we incorporate the domain knowledge of the control mechanism in the design of an action network to help the agent with action exploration. We test the algorithm with two challenging driving cases, the lane change situation and ramp merge situation.

The reminder of the paper is organized as follows. Related work is described in Section 2. Methodology is given in Section 3, followed by application case in Section 4 and experiments in Section 5. Conclusions and discussions are given in the last section.

\section{Related Work}
\label{related work}

In autonomous driving field, a vast majority of studies on operational level of driving are based on traditional methods. For example, in \cite{ho2009lane}, a virtual trajectory reference was created by a polynomial function for each moving vehicle, and a bicycle model was used to estimate vehicle positions based on the pre-calculated trajectories. In \cite{choi2015lane}, a number of way points obtained from Differential Global Positioning System and Real-time Kinematic devices were used generate a path for the autonomous vehicle to follow. Such approaches can work well in predefined situations or within the model limits, however, they have limited performance in unforeseeable driving conditions. 

In recent years, we have seen a lot of applications of RL on the automated driving domain \cite{yu2016deep} \cite{ngai2007automated}\cite{wang2017formulation}\cite{sallab2016end}. For example, Yu et al. \cite{yu2016deep} explored the application of Deep Q-Learning methods to the control of a simulated car in JavaScript Racer. They discretized the action space into nine actions and found that the vehicle agent can learn turning operations when there were no cars on the raceway but cannot perform well in obstacle avoidance. Ngai et al. \cite{ngai2007automated} put multiple goals in the RL framework (i.e. destination seeking and collision avoidance) to address the overtaking problem. They also converted continuous sensor values into discrete state-action pairs. In all of these applications, the action space was treated as discrete and few interactions with the surrounding environment were considered. Wang et al. \cite{wang2017formulation} proposed a RL framework for learning on-ramp merge behavior, where a Long Short Term Memory (LSTM) was used to learn internal states and a Deep Q-Network was used for deciding the optimal control policy.

Sallab et al. \cite{sallab2016end} moved further to explore the impacts of a discrete and a continuous action space on the lane keeping case. They conducted experiments in a simulated environment, and concluded that the vehicle agent traveled more smoothly under continuous action design than discrete action design.

Q-learning is simple but effective, and is basically applicable to discrete action space. If a Q-function approximator can be designed to encode continuous action values to corresponding Q-values, it becomes an optimization problems in a continuous action space \cite{wang2018reinforcement}. And it also avoids involving a complicated policy network as in most policy gradient based methods \cite{sutton2000policy}\cite{degris2012model}. Based on these thoughts, we design a quadratic Q-network, similar to the idea of Normalized Advantage Functions (NAF) \cite{gu2016continuous} in which the advantage term is parameterized as a quadratic function of nonlinear features of the state. We apply the method to the practical application case in autonomous driving, and combine it with domain knowledge of vehicle control mechanism to assist its action exploration. 

\section{Methodology}
\label{methodology}

\subsection{Quadratic Q-network}

In our RL formulation, the state space $S$ and the action space $A$ are taken as continuous. The goal of the reinforcement learning is to find an optimal policy $\pi^*:S\rightarrow A$, so that the total return $G$ accumulated over the course of the driving task is maximized. 
For a given policy $\pi$ with parameters $\theta$, a Q-value function is used to estimate the total reward from taking action $a_{t}$ given state $s_{t}$ at time $t$ \cite{sutton2000policy}. A value function is used to estimate the total reward from state $s_{t}$. The advantage function is to calculate how much better $a_{t}$ is in state $s_{t}$ as
\begin{equation}
\label{eq3}
A^\pi(s_{t}, a_{t})=Q^\pi(s_{t}, a_{t})-V^\pi(s_t).
\end{equation}


In the case where the action space is discrete, we can obtain the optimal action $a_t^*$ with the greedy policy directly by iterating over the action space, as
\begin{equation}
\label{eq55}
a_t^*=\mathrm{argmax}_{a_{t}} Q^\pi(s_{t}, a_{t}).
\end{equation}

However, when the action space is continuous, it is not easily ready to apply the basic Q-learning formula to find the optimal action. According to the nature of the quadratic equation, if the Q-function $Q^\pi(s_{t}, a_{t})$ has a quadratic form, the optimal action can be obtained analytically and easily. With this idea, we design the Q-function in a quadratic format as
\begin{equation}
\label{eq5}
Q^\pi(s_{t}, a_{t})=A^\pi(s_{t}, a_{t})+V^\pi(s_t)=(\mu(s_t)-a_t)^T M(s_t) (\mu(s_t)-a_t) + V(s_t),
\end{equation}
where $\mu(s_t)$ is a vector with the same dimension of the action, $M(s_t)$  is a negatively semi-definite matrix, and $V(s_t)$  is considered as a the value function with a scalar value as output. With this special form of Q-function, the optimal action $a_t^*$ can still be obtained in a greedy way as in equation (\ref{eq55}) but given by $\mu(s_t)$. 

Figure 1 (left) depicts the architecture of the quadratic Q-network. $M(s_t)$ and $V(s_t)$ are built separately with a single multilayer perceptron (MLP) with two hidden layers. In contrast, $\mu(s_t)$ consists of three MLPs that are combined in a special way. We call $\mu(s_t)$ the action network and give its details in the next subsection. $M(s_t)$, $V(s_t)$, and $\mu(s_t)$ are combined in the function $f_{Q}$ which is the equation (\ref{eq5}). 

\begin{figure}
  \centering
  \includegraphics[width=\textwidth,height=\textheight,keepaspectratio]{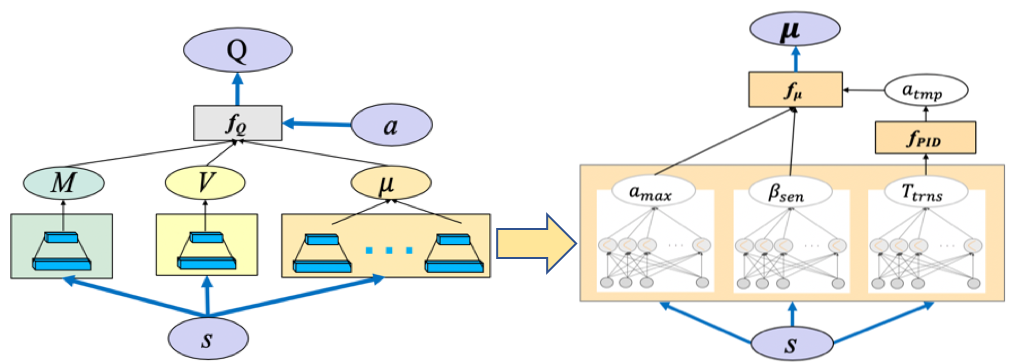}
  \label{fig1}
  \caption{Architecture of Quadratic Q-network (left) and action network $\mu$ (right).}
\end{figure}

Note that because any smooth Q-function should be Taylor-expanded in this quadratic format near the greedy action, there is not much loss in generality with this assumption if we stay close to the greedy policy that is being updated in the Q-learning process. 

\subsection{Action network }
From equation (\ref{eq5}) we can observe that $\mu(s_t)$ plays a critical role in learning the optimal action. If it is purely designed as a neural network with thousands of neurons, it may suffer a hard time learning actions meaningful to a driving policy. 

Based on this insight, we design the form of $\mu$ similar to a PID controller where some tuning parameters are replaced with neural networks. In other words, we do not manually tune the coefficients for the proportional, integral, and derivative terms but use neural networks to automatically find the appropriate values based on the defined reward function in RL. This way it makes the controller adaptable to different driving situations, and moreover the output action is based on a long-term goal of the task rather than an action just calculated for a target at current step as in a PID controller. 

The right graph in Figure 1 shows the design of action network $\mu(s_t)$ where three variables $a_{max}$, $\beta_{sen}$ and $T_{trs}$ are designed with neural networks, and Equation (\ref{eq6}) and (\ref{eq7}) show how these variables are combined. To be specific, from equation (\ref{eq6}), we obtain a temporary action based on PID properties, where $T_{trs}$ is the output from a neural network and interpreted as a transition time to mitigate errors between current and target states. The temporary value then goes through a hyperbolic tangent activation function in equation (\ref{eq7}), where another two parameters, $a_{max}$ and $\beta_{sen}$, are learned from neural networks. $a_{max}$ represents a tunable maximum acceleration and $\beta_{sen}$ indicates a sensitivity factor enforced on the temporary control action. Figure 1 (right) depicts the architecture of $\mu$.
\begin{equation}
    a_{tmp}=\frac{f(\Delta s)}{T^2_{trs}} + \frac{f'(\Delta s)}{T_{trs}}
    \label{eq6}
\end{equation}
\begin{equation}
    a=a_{max}*\mathrm{tanh}(\beta_{sen}*a_{tmp})
    \label{eq7}
\end{equation}
where $\Delta s$ is the state difference between a desired state and the current state. The desired state can be defined conveniently, for example, it can be the target lane ID for the lateral control or the preferred car-following distance for the longitudinal control. The state difference can include values such as relative distance $\Delta d$, relative speed $\Delta v$, and/or relative yaw angle $\Delta \phi$.

\subsection{Learning procedure}

There are two iterative loops in learning the policy. One is a simulation loop where it provides the environment that the vehicle agent interacts with, and the other one is a update loop in which the neural network weights are updated.

In the simulation loop, we use $\mu(s_t)$ to obtain the greedy action for a given state $s_t$ at step $t$. The greedy action is then perturbed with a Gaussian noise $n_t$ to increase its exploration and executed in the simulation. After the execution, we get a new state $s_{t+1}$ as well as a reward $r_t$ from the environment, and store the transition tuple ($s_t$, $a_t$, $s_{t+1}$,$r_t$) in a replay memory $D$. 

In the update loop, samples of tuples are drawn randomly from $D$. To overcome the inherent instability issues in Q-learning, we use experience replay technique and a target Q-network as proposed in \cite{adam2011experience}. Weights in Q-network ($\theta$) are updated by gradient descent at every time step, while weights in target Q-network ($\theta^-$) are periodically overwritten by $\theta$. Algorithm 1 gives the learning process. 


\begin{algorithm}
\caption{Quadratic Q-learning}
\label{alg:AIRL}
\setlength{\lineskip}{1pt}
\begin{algorithmic}[1]
\State Initialize Q-network weights $\theta$ in $M$, $V$, $\mu$
\State Initialize target Q-network weights $\theta^- \leftarrow \theta$
\For {episode $i=1$ to $N$}
\State Initialize action exploration noise $n_t \sim N$ 
\State Obtain initial state $s_0$
\For {$t=1$ to $T$}
\State Select an action $a_t=\mu_{\theta}(s_t) + n_t$
\State Execute the action $a_t$ and store the experience tuple ($s_t$, $a_t$, $s_{t+1}$,$r_t$) in $D$
\If {pretrain}
\State Update Q-function weights in $M$ and $V$ while keep $\mu$ frozen with samples from $D$ 
\EndIf 
\State Update all the weights $\theta$ in Q-function (\ref{eq5}) with batches drawn from $D$ for $m$ iterations
\State Update all the weights $\theta^-$ in target Q-network  periodically with $\theta$ from Q-network

\EndFor
\State \textbf{end for}
\EndFor
\State \textbf{end for}

\end{algorithmic}
\end{algorithm}

It is also worth mentioning that the overall training process includes two steps, pre-training step and training step. In pre-training, we only train the neural networks of $M$ and $V$, and froze the parameters in $\mu$. During training, we jointly update parameters in all neural network. This trick helps the agent learn faster. 

\subsection{Reward function}
In our study, the immediate reward is designed as a linear combination of multiple feature functions with respect to driving safety, comfort and efficiency. Each state-action pair is evaluated with a negative value, thus teaching the agent to avoid resulting in situations with large penalties.

To be specific, safety is evaluated by relative distances from vehicles that matter most to the ego vehicle. It includes relative distance to vehicles on the longitudinal direction and the distance to the center-line of the target lane on the lateral direction.  
\begin{equation}
    R_{s}=-w_{s1}(\sum_{l=1}^{L} f_{s1}(\Delta d_{lg})_l - w_{s2} f_{s2}(\Delta d_{lt})
    \label{eq9}
\end{equation}
where $R_{s}$ is the safety reward term, $f_{s1}$ and $f_{s2}$ are the feature functions which can be power functions based on how much we want to rate this feature, $w_{s1}$ and $w_{s2}$ are the weights, $\Delta d_{lg}$ and $\Delta d_{lt}$ are the relative distance on longitudinal and lateral directions, and $L$ is the number of adjacent vehicles.   

Comfort is evaluated by the control variables, $a_{lg}$ and $a_{lt}$ (i.e. speed acceleration and yaw acceleration), and their derivatives, $b_{lg}$ and $b_{lt}$.
\begin{equation}
    R_{c}=-w_{c1}(f_{c1}(a_{lg}, a_{lt}) - w_{c2}f_{c2}(b_{lg}, b_{lt})
    \label{eq10}
\end{equation}
where $ R_{c}$ is the comfort reward term, $f_{c1}$ and $f_{c2}$ are feature functions, and $w_{c1}$ and $w_{c2}$ are the weights.  

Efficiency is evaluated by the maneuvering time, i.e. how long it takes to finish the task. For example, in a merging case, it is the time consumed from the initiation to the completion of the behavior. The efficiency at a single time step is calculated by the time step interval ($\Delta t$).
\begin{equation}
    R_{t}=-w_{t}*f_{t}(\Delta t)
    \label{eq11}
\end{equation}
where $R_{t}$ is the efficiency reward term, $f_{t}$ is the feature function that can also be a power function of $\Delta t$, and $w_{t}$ is the function weight. 

Function weights are hyperparameters that are manually tuned through multiple training episodes. Their values and the expressions of the feature functions is given in the next section.

\section{Applications on lateral and longitudinal control}
\label{application}

We apply the proposed algorithm to two use cases, a lane-change situation and a ramp-merge situation. In the lane-change scenario, the lateral control is learned while the longitudinal control is an adapted Intelligent Driver Model (IDM) \cite{treiber2000congested}. In the ramp-merge scenario, the longitudinal control is learned while the lateral control is to follow the center-line of the current lane. We defer the work of simultaneously learning control variables on the two directions to our future work. 

There assumes to be a decision-making module and a gap selection module in the higher level that issue commands on when to make lane change or ramp merge. Our work focuses on learning the control variables under the received the commands. 


\subsection{Lateral control under lane-change case}
The lane change behavior is affected by the ego vehicle's kinematics (e.g. vehicle speed, position, yaw angle, yaw rate etc.) as well as the surrounding vehicles' in the targeting gap. Road curvature also affects the success of a lane change, for example, a curved road segment introduces additional centrifugal force that should be considered in the lane change process. Therefore, we define the state space to include both vehicle dynamics and road curvature information.

As mentioned earlier, we resort to a well-developed car-following model, Intelligent Driver Model (IDM) \cite{treiber2000congested}, with some adaptation for the longitudinal control. IDM describes dynamics of the positions and velocities of single vehicles. Due to space limitation, we only briefly introduce the modified IDM that is adapted to alleviate overly conservative driving behaviors. The longitudinal acceleration $a_{lg}$ is calculated by equation (\ref{eq13}). 
\begin{equation}
    a_{lg} = a_m(1-\mathrm{max}((\frac{v_{\alpha}}{v_0})^\delta, (\frac{s_0+v_{\alpha}T}{s_{\alpha}}+\frac{v_{\alpha} \Delta v_{\alpha}}{2\sqrt{a_m b} s_{\alpha}})^2))
    \label{eq13}
\end{equation}
where  $\Delta v_{\alpha}$ is the velocity difference between the ego vehicle $\alpha$ and its preceding vehicle $\alpha-1$, $v_0$ is the desired velocity of the ego vehicle in free traffic, $s_0$ is the minimum spacing to the leader, $s_{\alpha}$ is the current spacing, $T$ is the minimum headway to the leader, $a_m$ is the maximum acceleration, $b$ is the comfortable braking deceleration, and $\delta$ is the exponential parameter. In our test case, we set $s_0$ to $1m$, $T$ to $1s$, $a_m$ to $2.0 m/s^2$, $b$ to $1.5 m/s^2$, and $\delta$ to 4.

The action space for lateral control is treated as continuous to allow any reasonable real values being taken in the lane change process. Specifically, we define the lateral control action to be the yaw acceleration, $a_{lt}= \ddot \theta$ with the consideration that a finite yaw acceleration ensures the smoothness in steering, where $\theta$ is the yaw angle.

The reward function, composed of the three parts of safety, comfort and efficiency, is given in Table \ref{tab:tb1}. In the safety part, only reward from the lateral direction is considered in which $\Delta d_{lt}$ is the lateral deviation from current position to the center-line of the target lane. Safety in the longitudinal direction is taken care of by the gap selection and IDM model. The comfort part is evaluated by the lateral action $a_{lt}$ and the yaw rate $\omega$. The efficiency is evaluated by time-step intervals.

\subsection{Longitudinal control under ramp-merge case}
In the ramp-merging case, when the gap selection module finds a proper gap on the merging lane, the vehicle agent will try to merge into it by adjusting its longitudinal acceleration while keeping itself in the middle of the lane. The state space in such a situation includes the speed, position, heading angle of the ego vehicle, its leading vehicle and vehicles from the target gap. The action space is the longitudinal acceleration with continuous values in a limited range of $[-4.5 m/s^2, 2.5 m/s^2]$. 

The reward function is given in Table \ref{tab:tb1}. The safety term is decided by relative distances to the leading and lagging vehicles of the target gap on the merging lane. No lateral deviation is considered as discussed above. The longitudinal acceleration decides the comfort reward term. Efficiency is evaluated by time-step intervals, the same as in the lane-change case. 

\begin{table}
  \caption{Reward function design for lane change and ramp merge}
  \label{tab:tb1}
  \centering
  \begin{tabular}{llllll}
    \toprule
    \cmidrule(r){1-5}
    \textbf{Functions}      & Lane change        & Ramp merge                    &\textbf{Weights}         & Lane change     & Ramp merge  \\
    \midrule
    $f_{s1}$    &  None     & $\sum_{l=1}^{2} (\Delta d_{lng})_l$    & $w_{s1}$    & None            &0.01 \\
    \textbf{$f_{s2}$}    & $|\Delta d_{lt}|$  & None                         & $w_{s2}$    & 0.05            & None \\
    \textbf{$f_{c1}$}    & $|a_{lt}|$        & $|a_{lg}|$                   & $w_{c1}$    & 0.5             & 0.5\\
    \textbf{$f_{c2}$}    & $|\omega|$          & None                         & $w_{c2}$    & 2.0             & None\\
    \textbf{$f_{t}$}     & $|\Delta t|$        & $|\Delta t|$                   & $w_{t}$     & 0.05            & 0.05\\
    \bottomrule
  \end{tabular}
\end{table}

\section{Simulation and results}

\subsection{Simulation environment}
\begin{figure}
    \centering
    \includegraphics{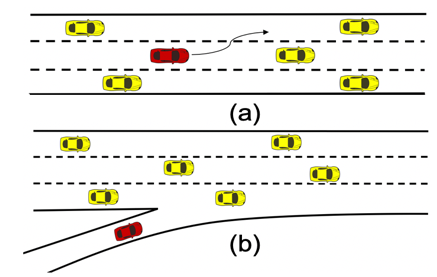}
    \caption{(a) Lane-change scenario (b) Highway ramp-merge scenario.}
    \label{fig3}
\end{figure}

The lane-change behavior is simulated on a highway segment of 350m long and three-lane wide on each direction. The ramp-merge behavior is simulated in the highway-ramp merging zone where the ramp is merged into the rightmost lane of the highway. An illustration of the simulation scene is show in Figure \ref{fig3}.

The simulated traffic is customized to generate diverse driving conditions. The initial speed, departure time interval, and speed limit of each individual vehicle is set to random values as long as they are within reasonable ranges, e.g. [30 km/h, 50 km/h], [5s, 10s], and [80 km/h, 120 km/h], respectively. In the simulation, vehicles can interact with each other. One example is that lagging vehicles in a lane-change case can yield or overtake the ego vehicle, creating diverse and realistic driving situations for training the RL agent.

The RL vehicle agent in the lane-change case is randomly generated in the middle lane thus that it can make either left or right lane change based on the command received after traveling for 150m. In the ramp-merge case, the RL agent is generated on the ramp, about 150m away from the merging intersection. Vehicles on the highway travel on its current lane and have the IDM car-following behavior. Additionally, a small portion of aggressive driving behaviors are simulated by setting a relatively high acceleration range and small car-following distances, and conversely for defensive driving behaviors. 

\subsection{Training results}
The hyperparameters for training in the two application cases are similar expect for the learning rate, 0.0005 for lane change and 0.001 for ramp merge, and training episodes, 6000 for lane change and 4000 for ramp merge. Other hyperparameters are set as follows: replay memory=2000, batch size=64, discount factor=0.95, target-Q weights update frequency=1000, optimizer=Adam. Twelve intermediate checkpoints are saved in each application case for testing the learned models.
	
\begin{figure}
    \centering
    \includegraphics[width=\textwidth,height=\textheight,keepaspectratio]{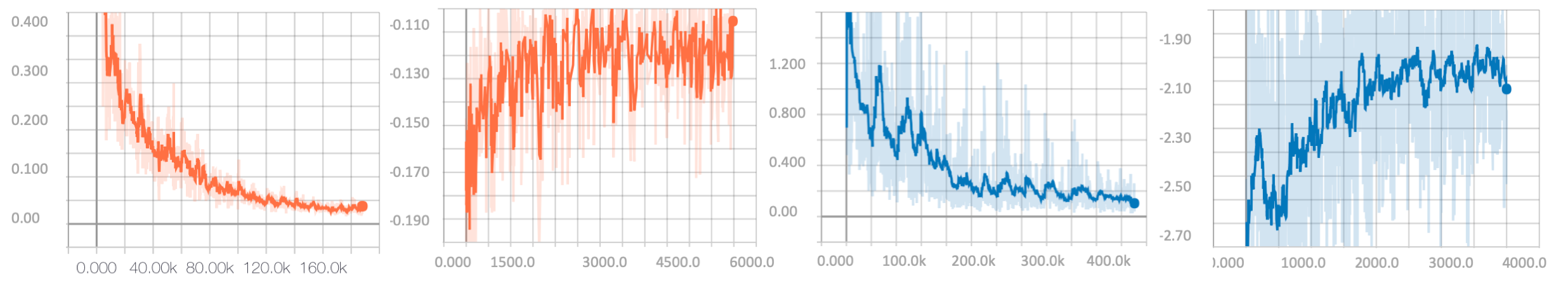}
    \caption{Training losses and accumulated rewards in lane-change case (in orange) and ramp-merge case (in blue) v.s. training steps, saved at different intervals. It shows that the loss decreases to a convergence and the accumulated reward increases to a convergence along with training steps.}
    \label{fig4}
\end{figure}

Training loss and accumulated rewards are plotted in Figure \ref{fig4} for both the lane-change case and ramp-merge case. 
From Figure \ref{fig4}, we can observe that in both cases the training loss curve shows an obvious convergence and that the total rewards also demonstrate a consistently increasing trend, which satisfactorily indicates that the RL vehicle agent has learned the lane-change behavior and ramp-merge behavior.  

\begin{figure}
    \centering
    \includegraphics[width=\textwidth,height=\textheight,keepaspectratio]{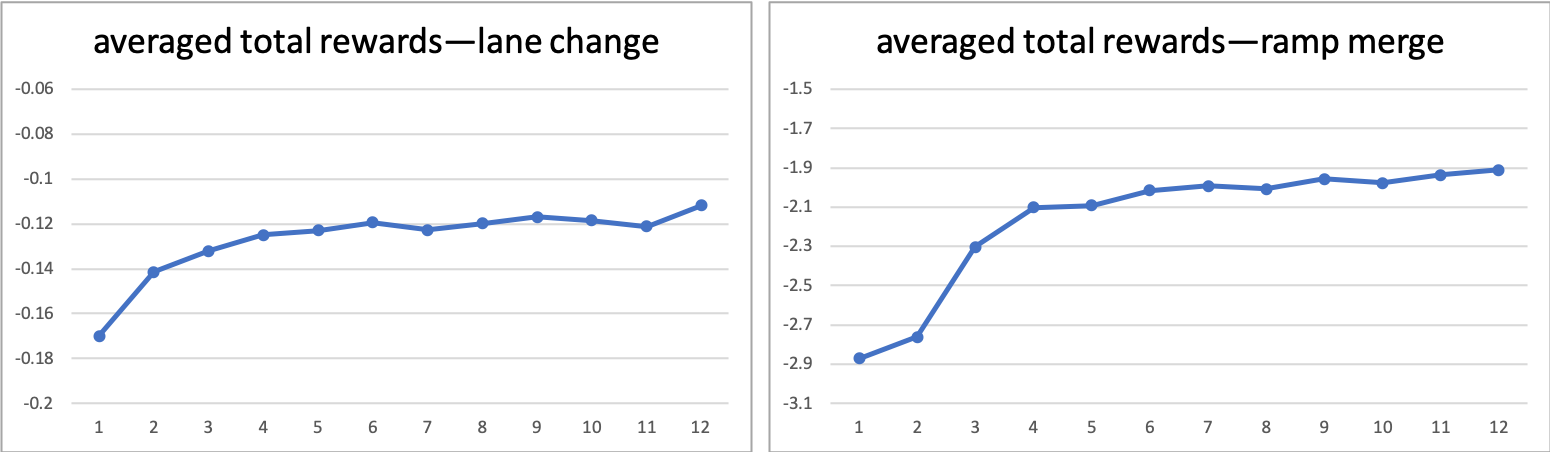}
    \caption{Testing performance. Averaged rewards over 100 test runs at each saved checkpoints v.s. saved checkpoints. Left: lane change case. Right: ramp merge case. It demonstrates that the averaged total rewards increase as training proceeds, consistent with the training results.}
    \label{fig5}
\end{figure}

Since each point in the total reward curves represents only one random driving case under that corresponding training step, it might not be enough to prove the learned driving behavior. Therefore, we conduct testing on the saved checkpoints to get an averaged driving performance. We run 100 episodes at each checkpoint in both the lane-change situation and ramp-merge situation, and then average their total rewards. The results are plot in Figure \ref{fig5}. 

The testing curves in Figure \ref{fig5} show consistent upward trend as the total reward curves in Figure \ref{fig4}, indicating that the RL agent has indeed progressively learned the driving behavior of lane change and ramp merge, and can take responsible actions with respect to safety, comfort and efficiency as defined in the reward function.

\begin{figure}
    \centering
    \includegraphics[width=\textwidth,height=\textheight,keepaspectratio]{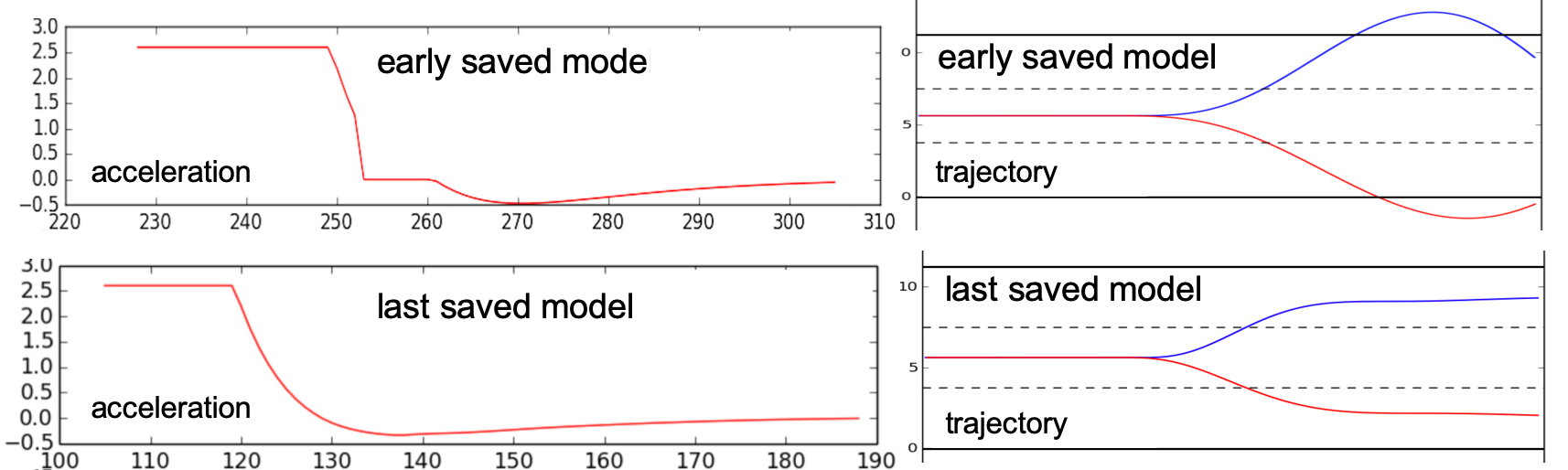}
    \caption{Acceleration comparison in ramp merge (left) and trajectory comparison in lane change (left). The graphs show that the acceleration and trajectory become more smooth and stable when training finishes. }
    \label{fig6}
\end{figure}

We also plot some driving dynamics to further compare the driving performance at the initial stage and the final stage, i.e., at the early saved checkpoint and last saved checkpoint. The right graph in Figure \ref{fig6} demonstrates the lane change trajectories (blue for left lane change and red for right lane change) at the initial stage (upper right) and the final stage (lower right), respectively. It shows clearly that the trajectories at the final stage, in comparison to initial stage, are quite smooth and stable. The left graph in Figure \ref{fig6} shows the acceleration curves in the ramp merge case as we learn the longitudinal control in this situation. We can see that the acceleration in a merging case gets smoother (lower left) in the final stage than that in the initial stage (upper left). 

\section{Conclusion and discussion}
In this work, we designed a Quadratic Q-network for handling continuous control problems in autonomous driving. With the quadratic format, the optimal action can be obtained easily and analytically. We also leverage domain knowledge of vehicle control mechanism for designing an action network, to provide the vehicle agent guidance in the action exploration. 

The proposed method is applied to two challenging driving cases, the lane-change case and ramp-merge case. Training results show  convergence in both the training losses and total rewards, indicating that the RL vehicle agent has learned to drive with higher rewards as defined in the reward function. Testing results show consistent convergence trend as that in the training, proving that the agent has indeed learned the behavior of lane changing and ramp merging. Comparison of the driving trajectories (in lane change situation) and vehicle accelerations (in ramp merge situation) at respectively the initial stage and final stage also reveals that the agent can drive safely, smoothly and efficiently. 

This study demonstrates the potentials of applying the quadratic    Q-learning framework to continuous control problems in autonomous driving. Our further step is to learn the longitudinal and lateral controls simultaneously based on different designs of reward functions. Also, we will try other directions for learning the policy, such as methods based on adversarial learning. Generative Adversarial Imitation Learning \cite{ho2016generative} and Adversarial Inverse Reinforcement Learning \cite{fu2017learning} show promising features in learning robotic control and it can recover both a policy and a reward function from demonstrations. Applying it to the dynamically changing environment in autonomous driving will be a challenging but interesting work.

\bibliographystyle{abbrv}
\bibliography{refs}
\end{document}